\documentclass[10pt,twocolumn,letterpaper]{article}

\usepackage{iccv}
\usepackage{times}
\usepackage{epsfig}
\usepackage{graphicx}
\usepackage{amsmath}
\usepackage{amssymb}
\usepackage{enumitem}
\usepackage{multirow}
\usepackage{xcolor}
\usepackage{subcaption}
\usepackage[precent]{overpic}
 
\usepackage[pagebackref=true,breaklinks=true,letterpaper=true,colorlinks,bookmarks=false]{hyperref}

\iccvfinalcopy % *** Uncomment this line for the final submission

\def\iccvPaperID{627} % *** Enter the ICCV Paper ID here

\ificcvfinal\fi
\begin{document}

%%%%%%%%% TITLE
\title{Mapped Convolutions}

\author{Marc Eder\quad\quad
True Price\quad\quad
Thanh Vu\quad\quad
Akash Bapat\quad\quad
Jan-Michael Frahm\\
University of North Carolina at Chapel Hill\\
Chapel Hill, NC\\
{\tt\small \{meder, jtprice, tvu, akash, jmf\}@cs.unc.edu}
}

\maketitle
%\thispagestyle{empty}

%%%%%%%%% ABSTRACT
\begin{abstract}
    We present a versatile formulation of the convolution operation that we term a ``mapped convolution.'' The standard convolution operation implicitly samples the pixel grid and computes a weighted sum. Our mapped convolution decouples these two components, freeing the operation from the confines of the image grid and allowing the kernel to process any type of structured data. As a test case, we demonstrate its use by applying it to dense inference on spherical data. We perform an in-depth study of existing spherical image convolution methods and propose an improved sampling method for equirectangular images. Then, we discuss the impact of data discretization when deriving a sampling function, highlighting drawbacks of the cube map representation for spherical data. Finally, we illustrate how mapped convolutions enable us to convolve directly on a mesh by projecting the spherical image onto a geodesic grid and training on the textured mesh. This method exceeds the state of the art for spherical depth estimation by nearly 17\%. Our findings suggest that mapped convolutions can be instrumental in expanding the application scope of convolutional neural networks.
\end{abstract}

% %%%%%%%%% INTRODUCTION
\section{Introduction}
For tasks on central perspective images, convolutional neural networks (CNN) have been a revolutionary innovation. However, the utility of these inference engines is limited for domains outside the regular Cartesian pixel grid. One of the most important properties behind the grid convolution's effectiveness, translational equivariance, can also be one of its most limiting factors. Translational equivariance with regards to discrete convolution means that the response of a filter convolved with a signal is unchanged by any shift of the signal. Mathematically this can be summarized as:
\begin{equation*}
    (f * g)[n] = (f * h(g))[n]
\end{equation*}

\begin{figure}[ht]
\begin{center}
\includegraphics[width=1\linewidth]{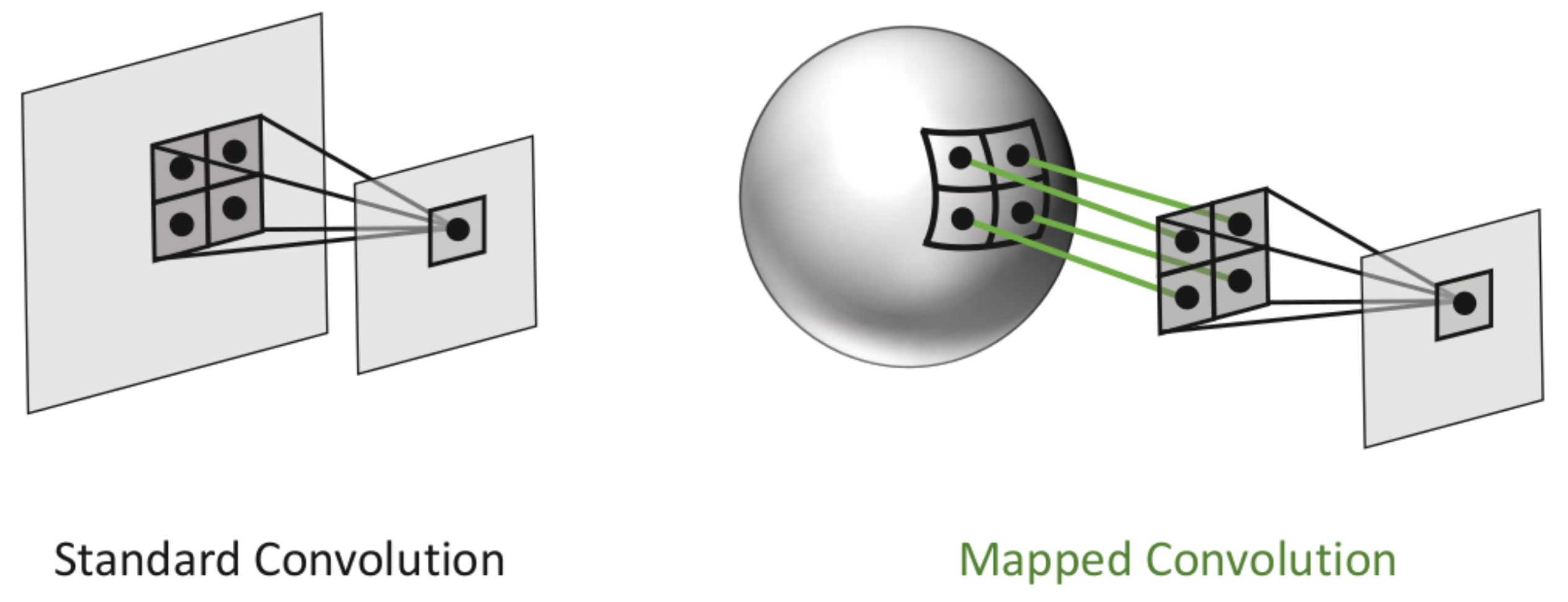}
\end{center}
   \vspace{-0.5cm}
   \caption{A visualization of how our proposed mapped convolution maps the grid sampling scheme to spherical data.}
   \vspace{-0.5cm}
\label{fig:teaser}
\end{figure}

\noindent for a filter $f$, a signal $g$, and a translation function $h$. In short, the filter response will be the same for a given set of pixels regardless of the set's location. This principle is what makes fully convolution networks possible \cite{long2015fully}, what drives the accuracy of state-of-the-art detection frameworks \cite{liu2016ssd}, and what allows CNNs to predict disparity and flow maps from stereo images \cite{mayer2016large}. Yet this property carries a subtler constraint: the convolutional filter and the signal must be discretized in a uniform manner. For standard 2D grid convolutions on images, this means the spacing between pixels must be regular across the image. When this constraint is broken, as can happen for data that is not organized on a regular grid, CNNs must accommodate for the irregularity.

There is a rich corpus of existing work looking to address this problem in the field of \textit{geometric deep learning} \cite{bronstein2017geometric}. For example, researchers have developed adaptive convolutions for specific distortion \cite{coors2018spherenet, tateno2018distortion}, extended convolutions to process non-Euclidean manifolds \cite{masci2015geodesic, monti2017geometric}, and leveraged information such as texture maps \cite{huang2018texturenet} to learn dense prediction tasks on 3D meshes. However, each of these methods requires a unique formulation of the convolution operation. 

We present a versatile formulation of the convolution operation that we call \textit{mapped convolution}, which provides a unifying interface for convolution on irregular data. This operation decouples the sampling function from the convolution operation, allowing popular CNN architectures to be applied to heterogeneous representations of structured data. The operation accepts a task- or domain-specific mapping function in the form of an adjacency list that dictates where the convolutional filters sample the input. This mapping function is the key to extending CNNs beyond the images to domains such as 3D meshes

In this work, we demonstrate the utility of mapped convolution operations by applying them to dense prediction on spherical images. With the growing availability of consumer-grade $180^\circ$ and $360^\circ$ cameras, spherical images are becoming an increasingly relevant domain for computer vision tasks. Furthermore, the spherical domain provides a useful test case for mapped convolutions. Because the images are represented on a sphere, we can represent them as planar projections (e.g. equirectangular images and cube maps) or as textures on a geodesic grid (e.g. icospheres). This allows us to demonstrate the utility of mapped convolutions in both the 2D and 3D realm while addressing a burgeoning new domain. We use these experiments to draw conclusions about designing useful mapping functions. 

We summarize our contributions as follows:
\begin{itemize}[topsep=0pt,itemsep=-1ex,partopsep=1ex,parsep=1ex]
    \item We introduce the mapped convolution operation, which provides an interface to adapt the convolution sampling function to irregularly-structured input data
    \item We demonstrate important considerations in designing a mapping function for an input domain by evaluating methods for dense depth estimation from spherical images
    \item Finally, we propose a mapping function to a geodesic grid that results in a nearly 17\% improvement over the state-of-the-art methods for spherical image depth estimation.
\end{itemize}

% %%%%%%%%% RELATED WORK
\section{Related Work}
\subsection{Data transformation}
Researchers have long been striving to augment the basic convolution operation for CNNs. Jaderberg \etal \cite{jaderberg2015spatial} introduced spatial transformer networks which allow the network to learn invariance to affine transformations of the inputs. This can be posed as a tool for visual attention, where the network learns to adjust the input according to the region(s) of interest. Similarly, Dai \etal \cite{dai2017deformable} developed the deformable convolution, which allows a CNN to learn an input-dependent sampling grid for the convolution operation. Jia \etal \cite{jia2016dynamic} dynamically learn the filters' operations themselves, conditioned on the inputs, which allows the network to be robust to local spatial transformations as well. Our mapped convolutions also address spatial transformations through a sampling function, but we are interested in the case where the transformation has a closed-form representation (e.g. the equirectangular distortion function) or can be defined (e.g. UV mapping from a texture to a 3D mesh), rather than regarding it as a latent variable to learn. This gives us an avenue to incorporate a prior knowledge of the structure of the data into training.

\subsection{Spherical imaging and geometric deep learning}
Spherical images have increased in popularity recently due to the growing accessibility of omnidirectional cameras and the benefits of the spherical domain's expanded field of view. Typically, these images are represented as equirectangular projections or cube maps. Although cube maps tend to suffer less from distortion, they are disjoint representations, with discontinuities at the edges of each face in the cube. Equirectangular projections preserve the spatial relationship of the content, and hence have been the more popular input domain for CNN-based methods to date. Due to the heavy distortion effects in equirectangular images, use of the traditional grid convolution causes performance to degrade significantly. To address this problem, Su and Grauman \cite{su2017learning} train a CNN to transfer existing perspective-projection-trained models to the equirectangular domain using a learnable adaptive convolutional kernel to match the outputs. Zioulis \etal \cite{zioulis2018omnidepth} explore the use of rectangular filter-banks, rather than square filters, to horizontally expand the network's receptive field to account for the equirectangular distortion. Another promising method is the spherical convolution derived by Cohen \etal \cite{cohen2017convolutional, cohen2018spherical}. Spherical convolutions address the nuances of spherical projections by filtering \textit{rotations} of the feature maps rather than \textit{translations}. However, the spherical convolution requires a specialized kernel formulated on the sphere. Our work follows most closely to that of Coors \etal \cite{coors2018spherenet} and Tateno \etal \cite{tateno2018distortion}. These works independently present a method to adapt the traditional convolutional kernel to the spherical distortion at a given location in the image. In this way, the network can be trained on perspective images and still perform effectively on spherical images. By accepting an arbitrary mapping function, our mapped convolutions extend this notion beyond distortion, permitting the representation of any structured data.

Although spherical images are typically represented on a planar image, they can be thought of as the texture of a genus-0 surface. Hence, we can consider the analysis of spherical images to sit alongside that of manifolds and graphs within the purview of \textit{geometric deep learning} \cite{bronstein2017geometric}. Kipf and Welling \cite{kipf2016semi} introduced the notion of a graph convolutional network (GCN), which has become a hallmark of geometric deep learning papers thanks to the flexibility of graphs to effectively represent a broad scope of data. They formulate the graph convolution as a ``neighborhood mixing'' operation between connected vertices of a graph by way of adjacency matrix multiplication. We also leverage a graphical representation to define the sampling operation. However, our mapped convolution more resembles the geodesic convolution neural network (GCNN) from Masci \etal \cite{masci2015geodesic}, which maintains the ``correlation with template'' notion of traditional CNNs. In that work, the authors apply the GCNN to Riemmanian manifolds, using local radial patches on the manifold to define the support of the convolution operation. As our mapped convolution provides an interface to specify how to sample the input, it has the capacity to be applied similarly to non-Euclidean domains.

% %%%%%%%%% MAPPED CONVOLUTIONS
\section{Mapped Convolutions}
Mapped convolutions generalize the standard convolution operation beyond the space of regular planar grids. In this section, we define the operation and explain the relevance of the sampling function to CNNs.

\textbf{Terminology}
We call our operation ``mapped convolution,'' because it \textit{maps} a convolution's sampling locations from a grid to different locations in the input. This mapping generalizes the notion of sampling implicit to convolution and is not restricted to a specific scheme. In this paper, we use the terms sampling function and mapping function interchangeably.

\subsection{Overview}
Equation (\ref{eq:stdconv}) gives the discrete convolution\footnote{Typically, the actual implementation uses the cross-correlation operation, which differs by the sign of the filter index.} formula, in which $K$ is the size of the kernel, $f[m]$ is the kernel weight at index $m$, and $n$ indexes the signal being convolved, $g[\cdot]$. For simplicity, we only state the 1D case, but generally, $m$ and $n$ can be $d$-dimensional tuples.

\begin{equation}\label{eq:stdconv}
    (f * g)[n] =  \sum_{m = -\lfloor \frac{K}{2} \rfloor}^{\lfloor \frac{K}{2} \rfloor} f[m]g[n-m].
\end{equation}

We define our mapped convolution on the observation that this operation is simply a sampling of the inputs followed by a weighted summation, rewriting Equation (\ref{eq:stdconv}) as:
\begin{equation}\label{eq:stdconvsampling}
    (f * g)[n] =  \sum_{m = -\lfloor \frac{K}{2} \rfloor}^{\lfloor \frac{K}{2} \rfloor} f[m] \left(\sum_{l=-\infty}^{\infty}g[l]\delta[l-(n-m)]\right)
\end{equation}
Mapped convolution decouples these two components. It is formulated similarly in Equation (\ref{eq:mapconv}), with the primary difference being that $n$ now indexes a mapping function, $\mathcal{M}$, instead of directly indexing the input. Note that the kernel shape is no longer predefined; the relationship between the kernel center and the output location is relegated to the mapping function.
\begin{equation}\label{eq:mapconv}
    (f * g)[n] =  \sum_{m = -\lfloor \frac{K}{2} \rfloor}^{\lfloor \frac{K}{2} \rfloor} f[m]D\left(g, \mathcal{M}[n-m]\right)
\end{equation}
The mapping function, given by Equation (\ref{eq:mapfunc}), maps the sampling location of the standard grid convolution, $x_{std} \in \mathbb{Z}$, to a new location, $x_{mapped} \in \mathbb{R}$. $D(g, x_{mapped})$ is an interpolation function dictates how to sample the signal at real-valued indices. In short, the mapping function defines how the input will be sampled at each location where the filter is applied.
\begin{equation}\label{eq:mapfunc}
  \mathcal{M}: x_{std} \rightarrow x_{mapped}
\end{equation}

This function can be domain- or data-dependent, and often changes layer-to-layer. An example of a domain-dependent mapping is the gnomonic projection function proposed by Coors \etal \cite{coors2018spherenet} and Tateno \etal \cite{tateno2018distortion} for processing equirectangular images. It depends only on the resolution of the image. Conversely, convolving on 3D meshes is topology-dependent and thus mappings may differ between inputs. The choice of interpolation methods is also task-dependent. In our experiments, we use nearest-neighbor and bilinear interpolation in the image domain and barycentric interpolation in the mesh domain, but the operation is not limited to these three interpolation functions.

\begin{table*}[ht]
\begin{center}
\small
\begin{tabular}{|c|c|c|c|c|c|c|c|c|}
\hline
\multicolumn{2}{|c|}{\textbf{Mapping Function}} & \textbf{AbsRel $\downarrow$} &\textbf{ SqRel $\downarrow$} &\textbf{ RMSLin $\downarrow$} & \textbf{RMSLog $\downarrow$} & $\mathbf{\delta<1.25} \uparrow$ & $\mathbf{\delta<1.25^2} \uparrow$ & $\mathbf{\delta<1.25^3} \uparrow$ \\
\hline\hline
\multicolumn{9}{|c|}{\textit{Experiments 1, 3}}\\
\hline
\parbox[t]{2mm}{\multirow{5}{*}{\rotatebox[origin=c]{90}{Omnidepth}}}
& Std. Grid & $0.1184$ & $0.0549$ & $0.3584$ & $0.1673$ & $0.8673$ & $0.9756$ & $0.9940$ \\
& Rect. Filter Bank \cite{zioulis2018omnidepth} & $0.1133$ & $0.0509$ & $0.3461$ & $0.1617$ & $0.8773$ & $\mathbf{0.9779}$ & $\mathbf{0.9946}$ \\
& Inv. Gnomonic \cite{coors2018spherenet, tateno2018distortion} & $0.1124$ & $0.0514$ & $0.3483$ & $0.1628$ & $0.8764$ & $0.9764$ & $0.9940$ \\
& Inv. Equirect. (ours) & $\mathbf{0.1104}$ & $\mathbf{0.0498}$ & $\mathbf{0.3417}$ & $\mathbf{0.1611}$ & $\mathbf{0.8785}$ & $0.9768$ & $0.9942$ \\\cline{2-9}
& ISEA (ours) & $\mathbf{0.0965}$ & $\mathbf{0.0371}$ & $\mathbf{0.2966}$ & $\mathbf{0.1413}$ & $\mathbf{0.9068}$ & $\mathbf{0.9854}$ & $\mathbf{0.9967}$ \\
\hline\hline
\parbox[t]{2mm}{\multirow{5}{*}{\rotatebox[origin=c]{90}{SUMO}}}
& Std. Grid & $0.0842$ & $0.0616$ & $0.3655$ & $0.1729$ & $0.9131$ & $0.9642$ & $0.9844$ \\
& Rect. Filter Bank \cite{zioulis2018omnidepth} & $0.0792$ & $0.0553$ & $0.3535$ & $0.1658$ & $0.9197$ & $0.9624$ & $0.9834$ \\
& Inv. Gnomonic \cite{coors2018spherenet, tateno2018distortion} & $0.0767$ & $0.0573$ & $0.3534$ & $0.1660$ & $0.9224$ & $0.9670$ & $0.9853$ \\
& Inv. Equirect. (ours) & $\mathbf{0.0745}$ & $\mathbf{0.0534}$ & $\mathbf{0.3481}$ & $\mathbf{0.1627}$ & $\mathbf{0.9234}$ & $\mathbf{0.9682}$ & $\mathbf{0.9860}$ \\\cline{2-9}
& ISEA (ours) & $\mathbf{0.0628}$ & $\mathbf{0.0426}$ & $\mathbf{0.3084}$ & $\mathbf{0.1449}$ & $\mathbf{0.9422}$ & $\mathbf{0.9756}$ & $\mathbf{0.9893}$ \\
\hline\hline
\multicolumn{9}{|c|}{\textit{Experiment 2}}\\
\hline
\multicolumn{2}{|c|}{Cube Map (dim: $128$px)} & $0.0956$ & $0.0728$ & $0.3932$ & $0.1837$ & $0.9011$ & $0.9608$ & $0.9826$ \\
\multicolumn{2}{|c|}{Cube Map (dim: $256$px)} & $0.1256$ & $0.1053$ & $0.4586$ & $0.2210$ & $0.8546$ & $0.9409$ & $0.9743$ \\
\hline
\end{tabular}
\end{center}
\vspace{-3mm}
\caption{Quantitative results for depth estimation experiments on the SUMO \cite{sumo} and Omnidepth \cite{zioulis2018omnidepth} datasets. \textit{Experiment 1:} [Section \ref{sec:mapproj}] Comparing different sampling functions on equirectangular images. Our inverse equirectangular mapping function, defined in Equation (\ref{eq:revequirect}), clearly outperforms the other methods. \textit{Experiment 2:} [Section \ref{sec:cubemaps}] Cube map results using the mapping function given in Equation (\ref{eq:cubemapping}). \textit{Experiment 3:} [Section \ref{sec:icosphere}] Results when applying our proposed ISEA method. This approach significantly outperforms the existing state-of-the-art methods. \textit{Experiment 2} is performed on SUMO only.}
\vspace{-2mm}
\label{tab:mapprojresults}
\end{table*}

\subsection{Implementation details}\label{sec:implementation}
The standard convolution operation is typically implemented using some variant of the \textit{im2col} algorithm. In short, this approach samples a multi-dimensional tensor, copies it into a large matrix, and then leverages highly-parallelizable general matrix multiplication (GEMM) operations to apply the weights. We refer the interested reader to Anderson \etal \cite{anderson2017low} for an in-depth overview of convolution implementation variants and their efficiencies. For our mapped convolutions, we modify this core algorithm to accept an adjacency list defining the desired sampling function. This data structure is useful as it is compact but can represent any type of graphically-structured data. This enables convolution on non-Euclidean data.

\subsection{Relationship to existing methods}
The mapped convolution is a versatile operation that can be applied to a variety of different data domains. For many of these domains, specific operations have previously been developed. Our mapped convolution subsumes some of these previous methods; they can be seen as a special case of our general mapped convolution.

Consider the gnomonic convolution operation developed by Coors \etal \cite{coors2018spherenet} and Tateno \etal \cite{tateno2018distortion} for processing equirectangular projections. This domain-specific mapping function uses the inverse gnomonic projection function to convert Cartesian to spherical coordinates, $M: (x,y) \rightarrow (\phi, \psi)$. In Section \ref{sec:mapproj}, we evaluate this mapping function for equirectangular images and suggest an improvement based on the inverse equirectangular projection.

The geodesic convolution operation proposed by Masci \etal \cite{masci2015geodesic} is another application of mapped convolutions. The authors define a radial patch operator to sample from a Riemannian manifold based on the mapping function $M: (\rho, \theta) \rightarrow B(x)$, where $\rho$ and $\theta$ are local geodesic polar coordinates and $B(x)$ is the manifold of the mesh around some location $x$. In Section \ref{sec:icosphere}, we apply the mapped convolution to a simple geodesic, a subdivided icosahedron, using the inverse gnomonic projection function as the mapping.

Our proposed operation also shares some characteristics with graph convolution networks (GCN) from Kipf and Welling \cite{kipf2016semi}. However, there are notable distinctions between mapped convolutions and GCNs. Graph convolutions typically learn to embed nodes of a graph at each layer by performing a ``neighborhood-mixing'' of the data at adjacent nodes through multiplication with an adjacency matrix. In contrast, we maintain the ``correlation with template'' model of a sliding window kernel. There are pros and cons to both methods. For one, our method maintains a correspondence to grid convolutions which permits the extension of popular network architectures to irregularly structured data. On the other hand, we must use a constant kernel size to do this, which limits mapped convolution operations to graphs with uniform vertex degree. GCNs have no such restrictions on vertex degree, but they lack the transferability between domains that we maintain via analogy to grid convolution.

Finally, the deformable convolution introduced by Dai \etal \cite{dai2017deformable} can be viewed as an application where the mapping function is learned as parameter of the network rather than fixed as an additional input. However, this is not the goal of mapped convolutions. Rather, we aim to incorporate the known structure of the data into how we process it.

It is worth noting as well that the traditional grid convolution and its common library of parameters (e.g. stride, dilation, padding, etc.) can be expressed as a mapping function as well and can therefore be implemented as a mapped convolution.

\begin{figure*}[ht]
\begin{center}
\includegraphics[width=0.9\linewidth]{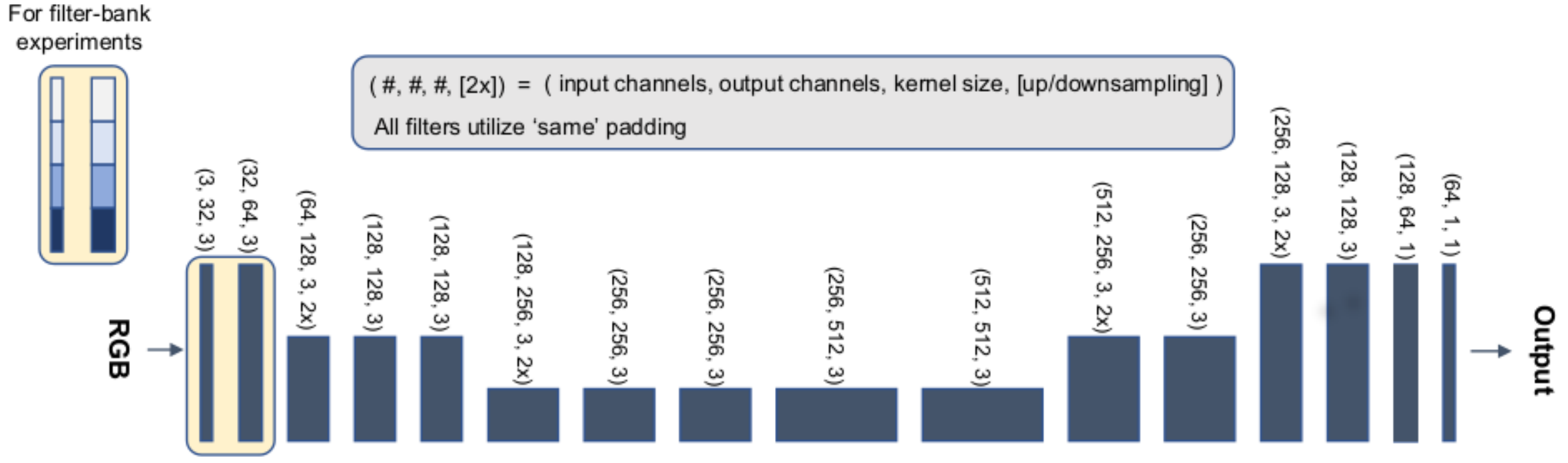}
\end{center}
\vspace{-2mm}
\caption{Simple encoder-decoder network architecture used for all experiments. For the filter bank method evaluated in Section \ref{sec:mapproj}, we split the first two layers channel-wise into 4 blocks, one for each filter, and concatenate the results as in \cite{zioulis2018omnidepth}.}
\label{fig:network}
\end{figure*}

% %%%%%%%%% EXPERIMENTS
\section{Experiments}\label{sec:results}
We illustrate the usefulness of the mapped convolution and the importance of the sampling function by applying mapped convolutions to dense depth estimation from spherical images. First, we show the importance of selecting the correct mapping function for the data domain. We analyze existing state-of-the-art methods and demonstrate that the inverse equirectangular projection is the ideal mapping function for this domain. Next, we explore the cube map representation of spherical images and empirically show the pitfalls of overlooking the discretization of the data in designing a mapping function. Finally, we address an overlooked issue of spherical images, information imbalance in the data representation. We suggest that a mapping function based on an icospherical geodesic grid resolves both this imbalance and the problem of spherical distortion. Using this mapping function and applying convolution to the vertices of the geodesic, we achieve a nearly 17\% improvement over the existing state-of-the-art methods in terms of absolute error.

For all experiments, we use the simple encoder-decoder network architecture shown in Figure \ref{fig:network}. In each trial, we train for 10 epochs using Adam optimization \cite{kingma2014adam} with an initial learning rate of $10^{-4}$, reduced by half every 3 epochs. We use the robust BerHu loss \cite{laina2016deeper} as our training criterion.

We use two large-scale spherical image datasets to evaluate our methods: Omnidepth \cite{zioulis2018omnidepth} and SUMO \cite{sumo}. The Omnidepth \cite{zioulis2018omnidepth} dataset is an aggregation of four existing omnidirectional image datasets, two consisting of real scene captures, Matterport3D \cite{Matterport3D} and Stanford 2D-3D \cite{stanford2d3d}, and two with exclusively computer-generated scenes, SceneNet \cite{scenenet} and SunCG \cite{suncg}. SUMO is a derivative of SunCG consisting of exclusively computer-generated images.

\subsection{Convolution on equirectangular images}\label{sec:mapproj}
Equirectangular projections are a common spherical image representation thanks to the simple association between pixel and spherical coordinates. Unfortunately, this relationship causes heavy distortion towards the poles of the sphere. Recent methods have tried to account for this distortion for dense prediction tasks. Zioulis \etal \cite{zioulis2018omnidepth} use rectangular filter-banks on the input layers of the network to address the strong horizontal distortion that occurs near the poles in the image, while Coors \etal \cite{coors2018spherenet} and  Tateno \etal \cite{tateno2018distortion} both sample from the image according to an inverse gnomonic projection. Both methods improve over standard square grid convolutions, but neither method completely accounts for the distortion induced by the equirectangular image. While the rectangular filter bank increases the horizontal receptive field of the filters, it does not attempt to model to the distortion. The approach of \cite{coors2018spherenet, tateno2018distortion} does address spherical distortion, but it does not apply the optimal model for equirectangular images. The gnomonic projection maps the sphere onto a tangent plane, which is a useful analogy for applying a grid convolution to a sphere. However, equirectangular projections are a different class of projection, mapping the sphere to a cylinder. Their distortion is different from the deformation induced by the gnomonic projection. As the image is formed via equirectangular projection, it is logical that sampling according to the \textit{inverse equirectangular projection} will `undo' the distortion. 
\begin{figure*}
    \centering
    \begin{subfigure}[b]{0.23\textwidth}
        \centering
        \includegraphics[width=\textwidth, trim={0.5cm 1.5cm 1.5cm 2.7cm}, clip] {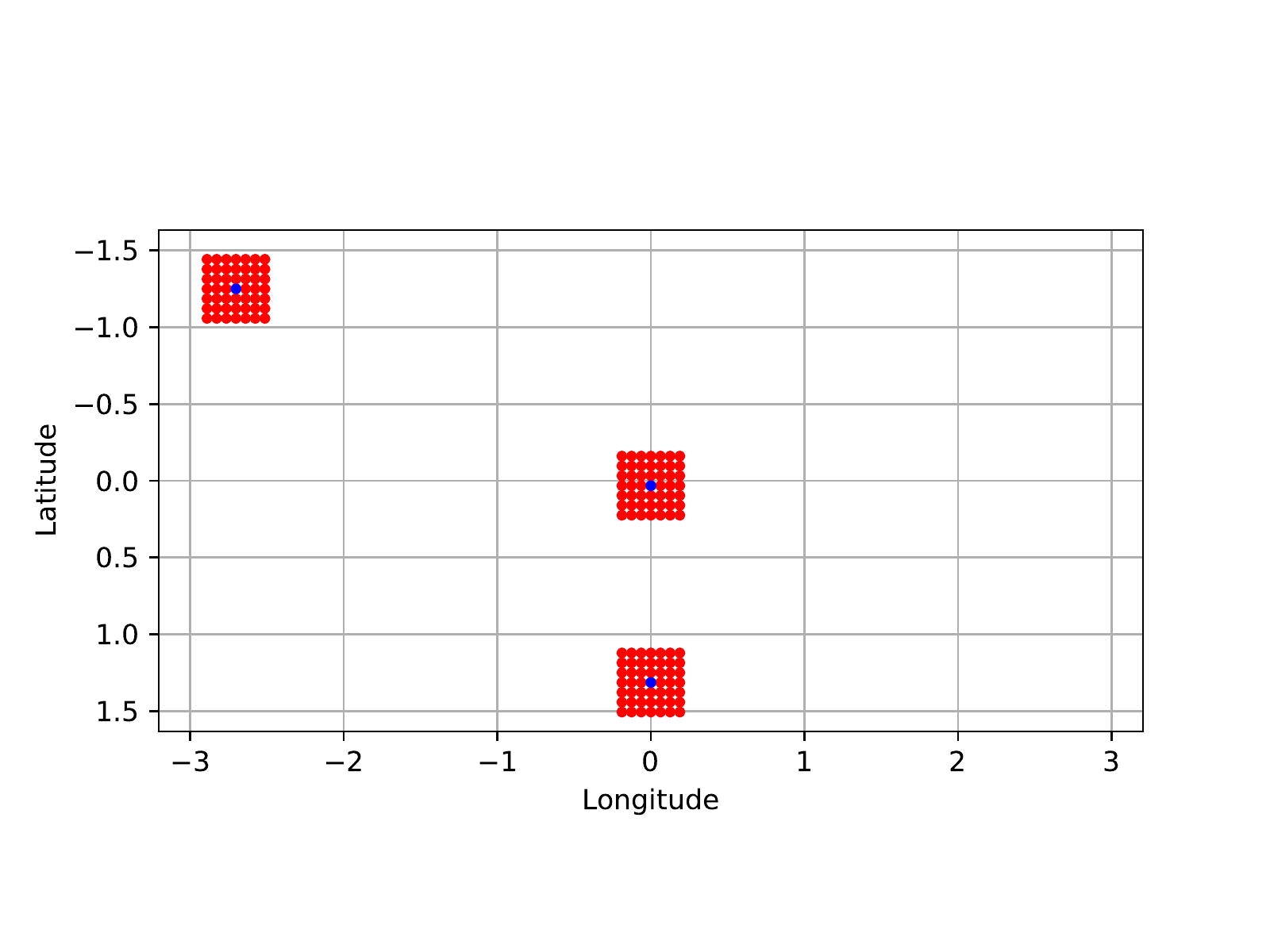}
        \caption{\label{fig:gridsampling}Std. Grid}
    \end{subfigure}
    ~
    \centering
        \begin{subfigure}[b]{0.23\textwidth}
        \centering
        \includegraphics[width=\textwidth, trim={0.5cm 1.5cm 1.5cm 2.7cm}, clip] {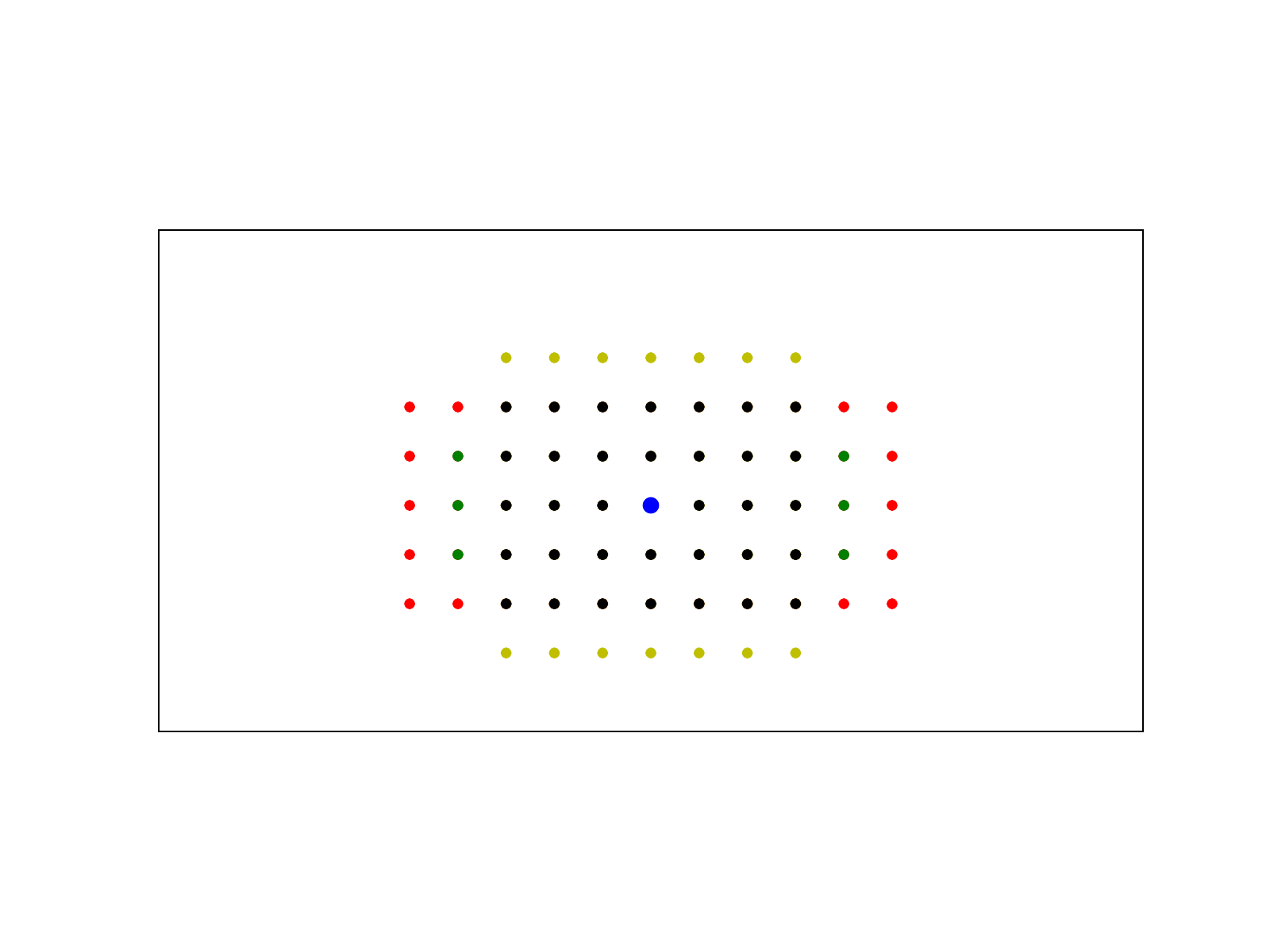}
        \caption{\label{fig:filterbanksampling}Rect. Filter Bank \cite{zioulis2018omnidepth}}
    \end{subfigure}
    ~
    \begin{subfigure}[b]{0.23\textwidth}
        \centering
        \includegraphics[width=\textwidth,trim={0.5cm 1.5cm 1.5cm 2.7cm}, clip] {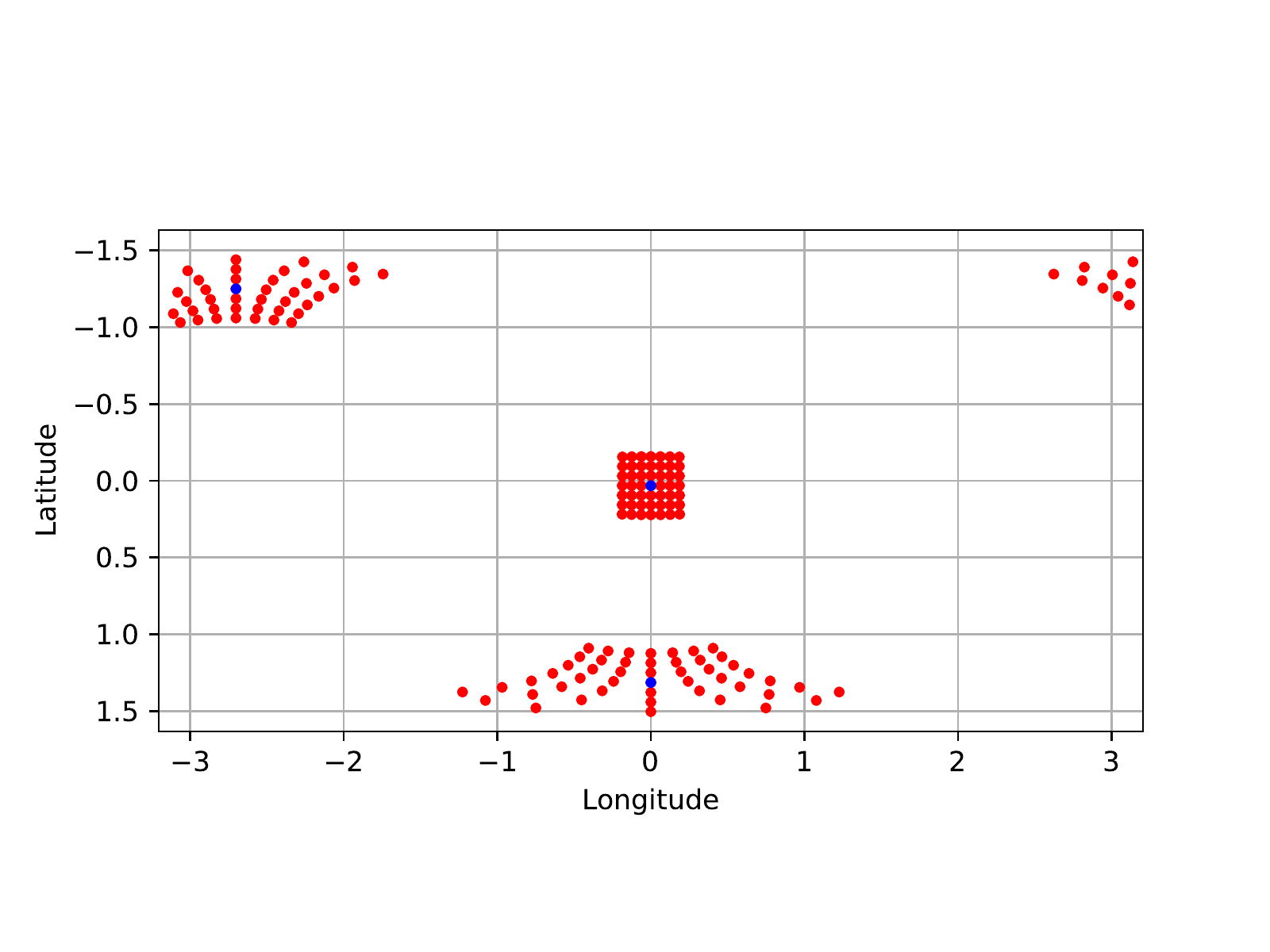}
        \caption{\label{fig:gnomonicsampling}Inv. Gnomonic \cite{coors2018spherenet, tateno2018distortion}}
    \end{subfigure}
    ~
    \begin{subfigure}[b]{0.23\textwidth}
        \centering
        \includegraphics[width=\textwidth, trim={0.5cm 1.5cm 1.5cm 2.7cm}, clip] {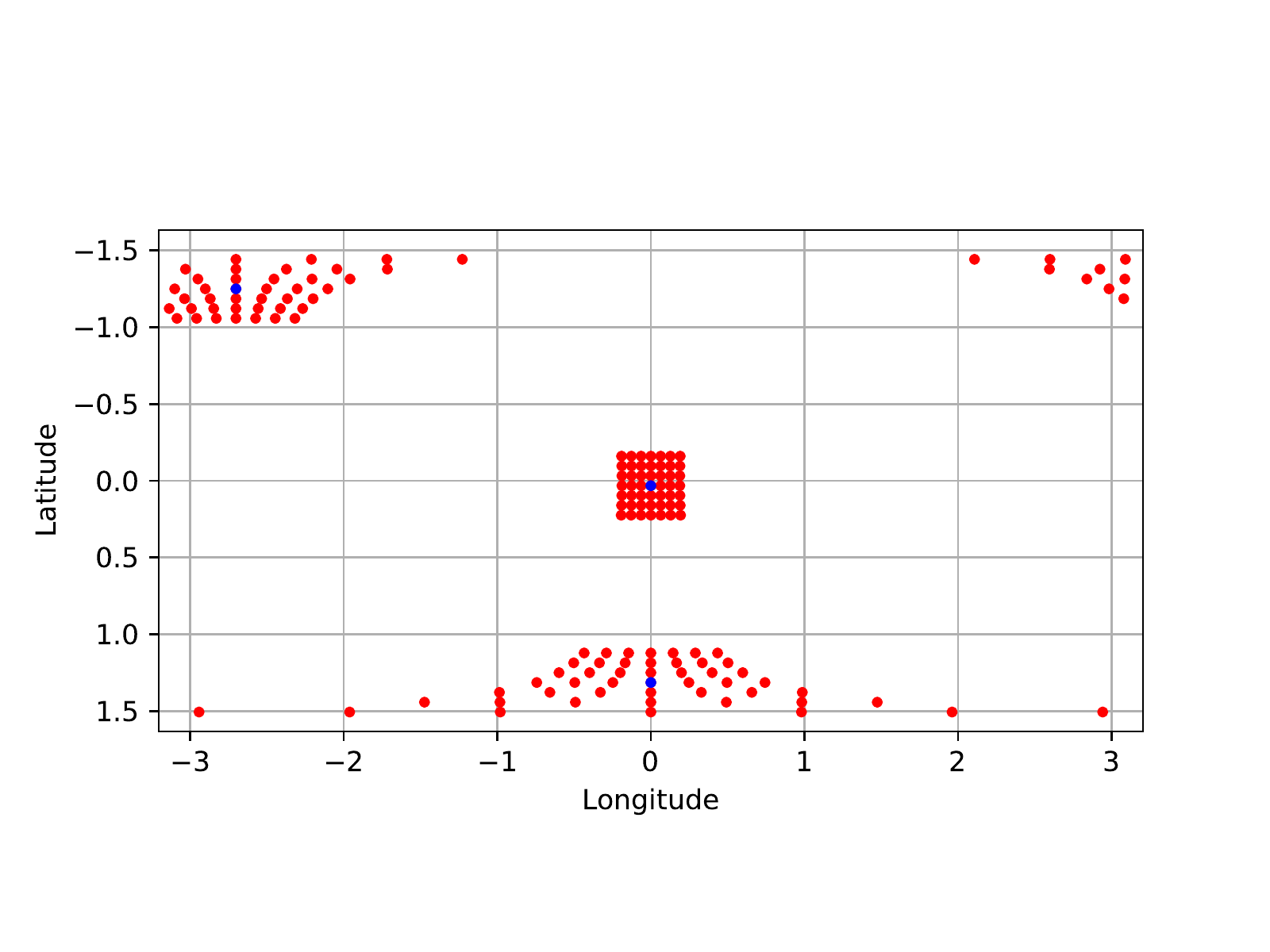}
        \caption{\label{fig:equirectsampling}Inv. Equirect. (ours)}
    \end{subfigure}
\vspace{-2mm}
\caption{A comparison of sampling functions used for spherical images. The blue points represent the kernel center. Note how the inverse equirectangular approach preserves samples along lines of latitude and stretches much wider toward the poles, while the gnomonic sampling curves lines of latitude and covers a smaller region.}
\vspace{-2mm}
\end{figure*}

This is borne out by the projection functions themselves. The following defines the inverse gnomonic projection function (i.e. image-to-sphere):
\begin{align}\label{eq:invgnomonic}
    &\phi = \sin^{-1} \left(\cos c \sin y_{0} + \frac{\Delta y_{(i,j)} \sin c \cos y_{0}}{\rho} \right) \\
    &\lambda = x_{0} + \tan^{-1} \left( \frac{\Delta x_{(i,j)} \sin c}{\rho \cos y_{0} \cos c - \Delta y_{(i,j)} \sin y_{0} \sin c} \right) \notag \\
    &\rho = \sqrt{\Delta x_{(i,j)}^2 + \Delta y_{(i,j)}^2} \;\;\;\; c = \tan^{-1}\rho, \notag
\end{align}
where $\phi$ and $\lambda$ represent latitude and longitude on the sphere, and $\Delta x_{(i,j)}$ and $\Delta y_{(i,j)}$ define the angular distance in the $X$ and $Y$ directions, respectively, from the kernel center $(x_{0}, y_{0})$ at kernel index $(i,j)$.

Notice that the resulting spherical coordinates are both functions of the $x$ and $y$ coordinates in the image. This relationship is visible in Figure \ref{fig:gnomonicsampling}, which shows the inverse gnomonic sampling function. The line of points sampled near the poles visibly curves. On the other hand, equirectangular projections map parallels of latitude directly to rows in an image. In other words, the $y$ coordinate in the image should only be a function of the latitude. Indeed, this is the case with the inverse equirectangular projection:%
\vspace{-1mm}
\begin{align}\label{eq:revequirect}
    &\phi = y_{0} + \Delta y_{(i,j)} \\
    & \lambda = x_{0} + \Delta x_{(i,j)} \sec \phi \notag
\end{align}
Figure \ref{fig:equirectsampling} shows samples generated using the inverse equirectangular mapping. While similar to the inverse gnomonic sampling, the points noticeably spread wider to accommodate the distortion near the poles, and rows of the sample grid project to rows in the image as well. This difference is slight, but we show that it is an important nuance.%

We compare the inverse equirectangular projection mapping to the inverse gnomonic projection \cite{coors2018spherenet, tateno2018distortion}, the rectangular filter-bank method \cite{zioulis2018omnidepth}, and the standard square grid convolution. We train each convolution variant three times on each dataset and report the average results for the standard set of depth estimation metrics in experiments 1 in Table \ref{tab:mapprojresults}. While all three methods outperform the standard grid convolution, the results clearly show that our proposed inverse equirectangular mapping surpasses the existing methods. This outcome highlights the importance of selecting the appropriate mapping function for the data domain.

\begin{figure*}
    \centering
    \begin{minipage}{.94\linewidth}
        \centering
        \begin{subfigure}[b]{0.32\textwidth}
            \includegraphics[width=\textwidth]{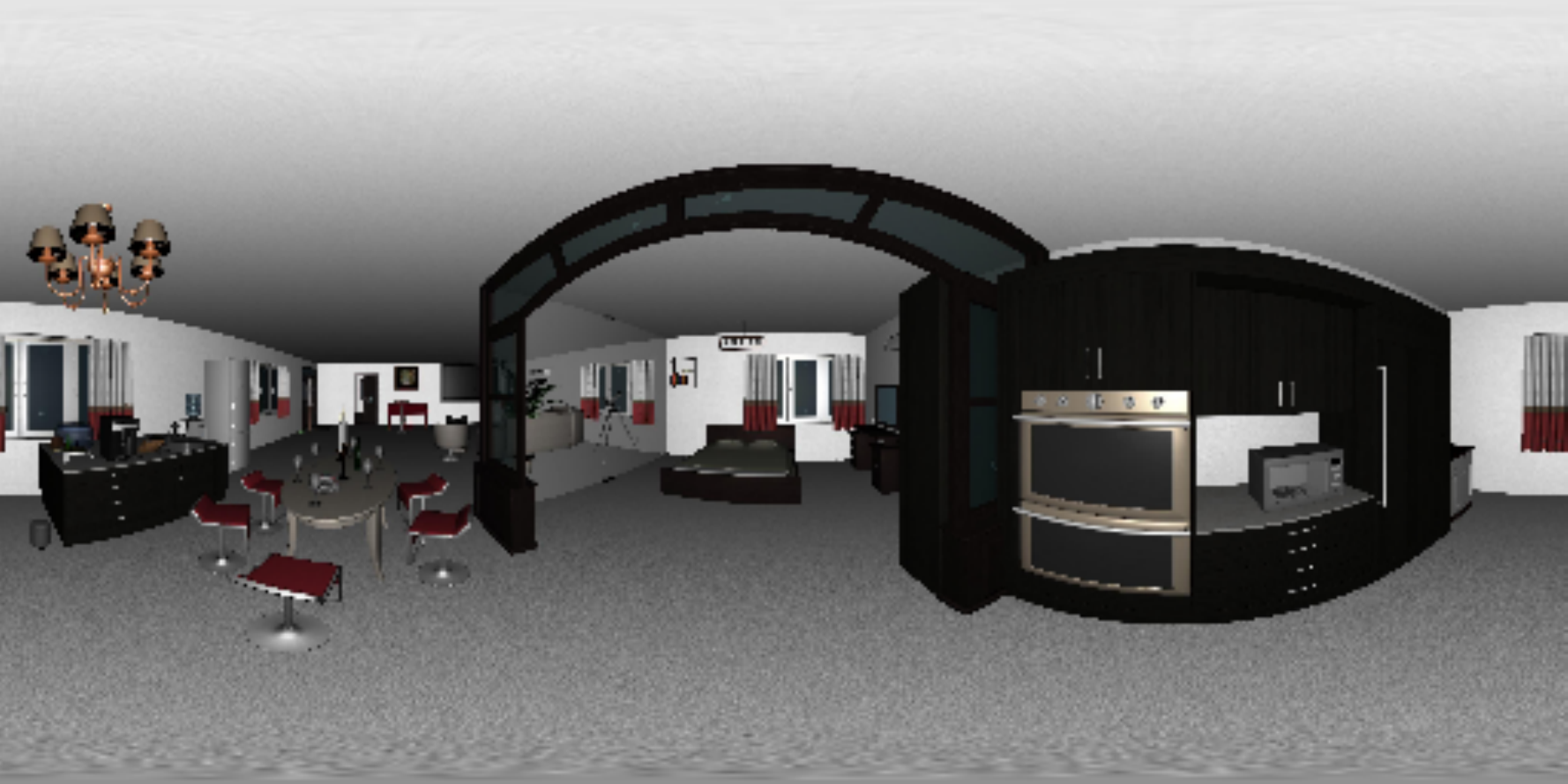}
        \end{subfigure}
        ~
        \centering
        \begin{subfigure}[b]{0.32\textwidth}
            \begin{overpic}[width=\textwidth]{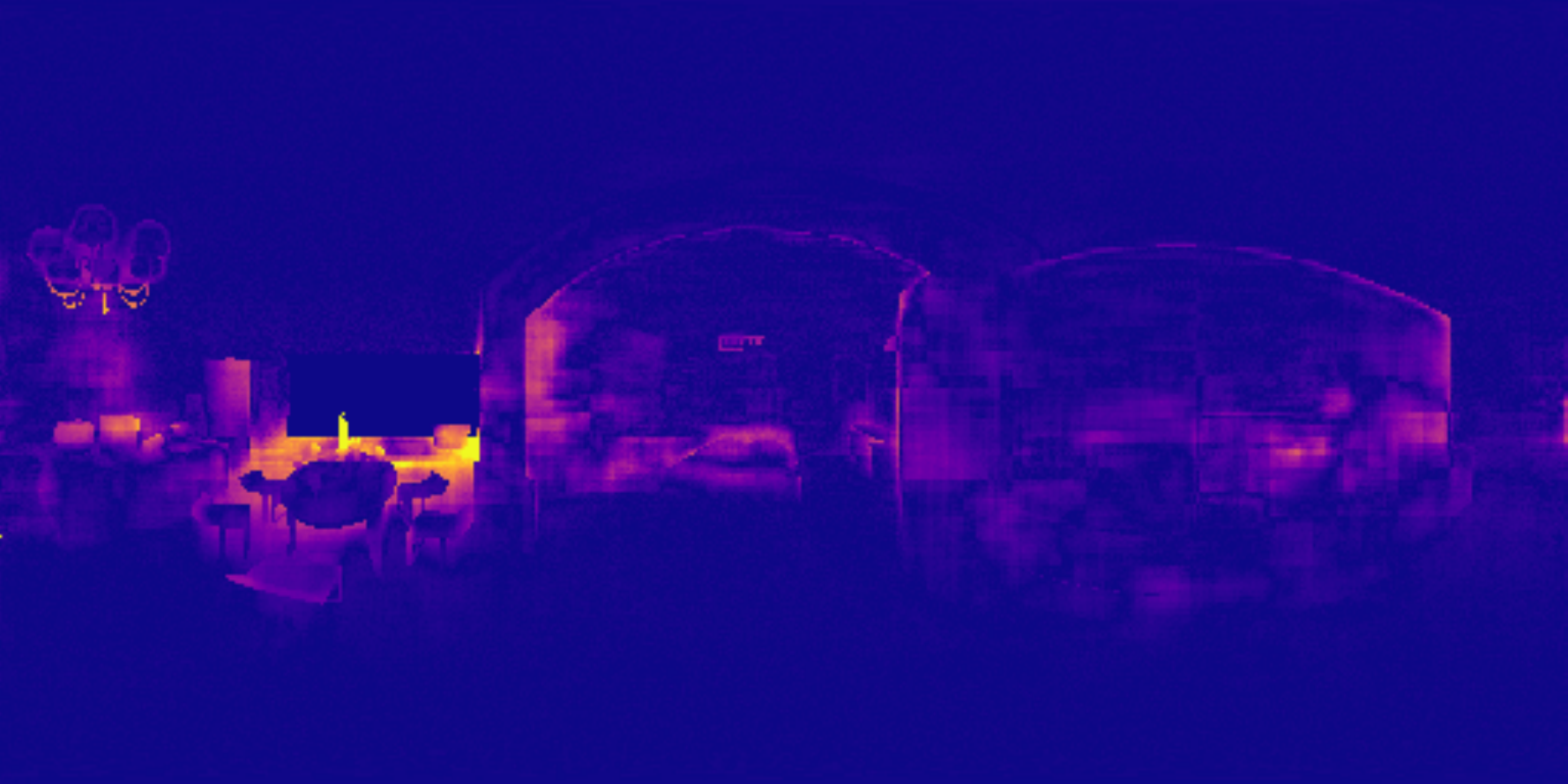}\put(2,40){\textcolor{white}{\textsf{Std. Grid}}}\end{overpic}
        \end{subfigure}
        ~
        \centering
        \begin{subfigure}[b]{0.32\textwidth}
            \begin{overpic}[width=\textwidth]{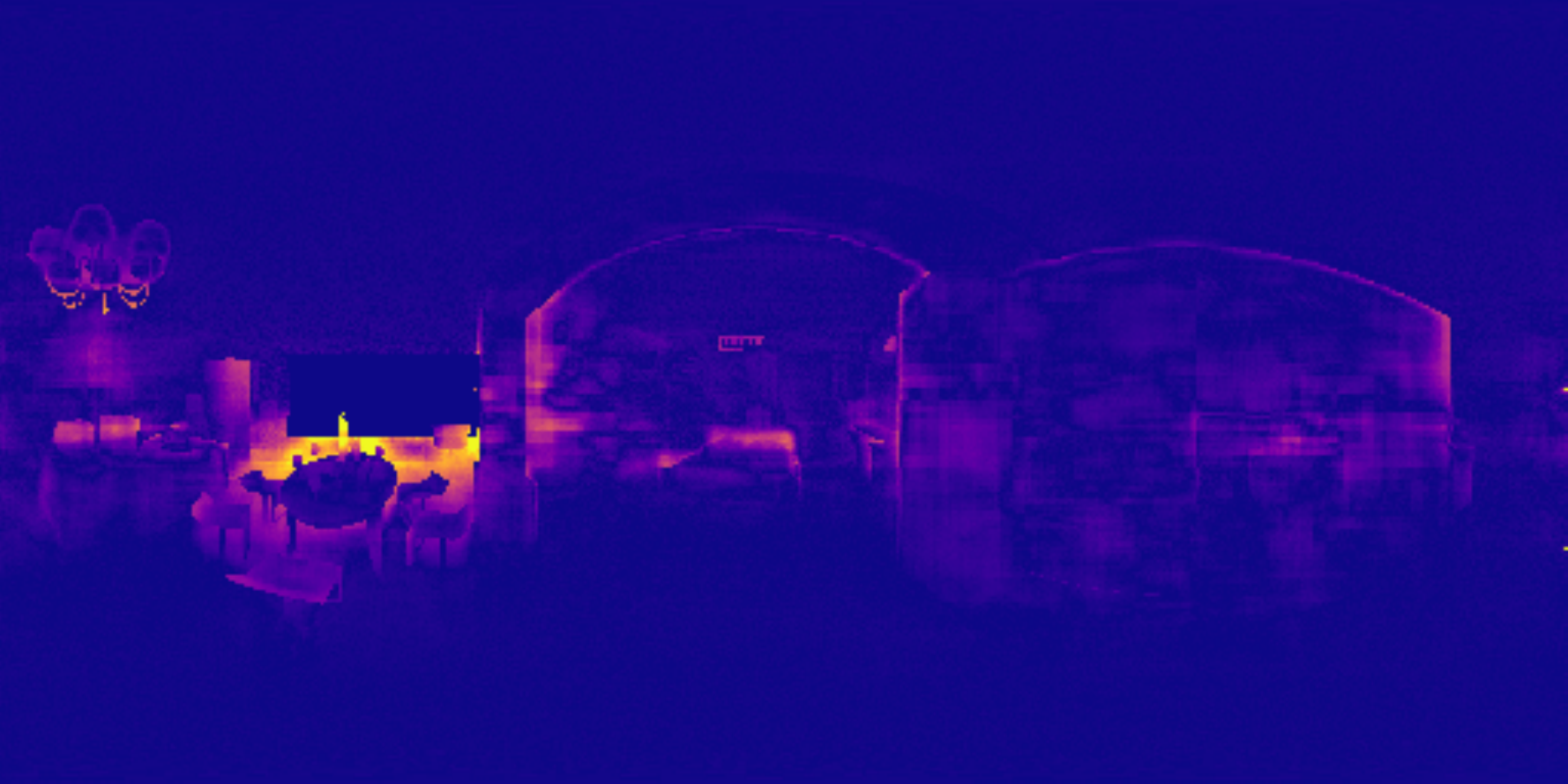}\put(2,40){\textcolor{white}{\textsf{Rect. Filter Bank \cite{zioulis2018omnidepth}}}}\end{overpic}
        \end{subfigure}\\
    \vspace{4pt}
        \centering
        \begin{subfigure}[b]{0.32\textwidth}
            \begin{overpic}[width=\textwidth]{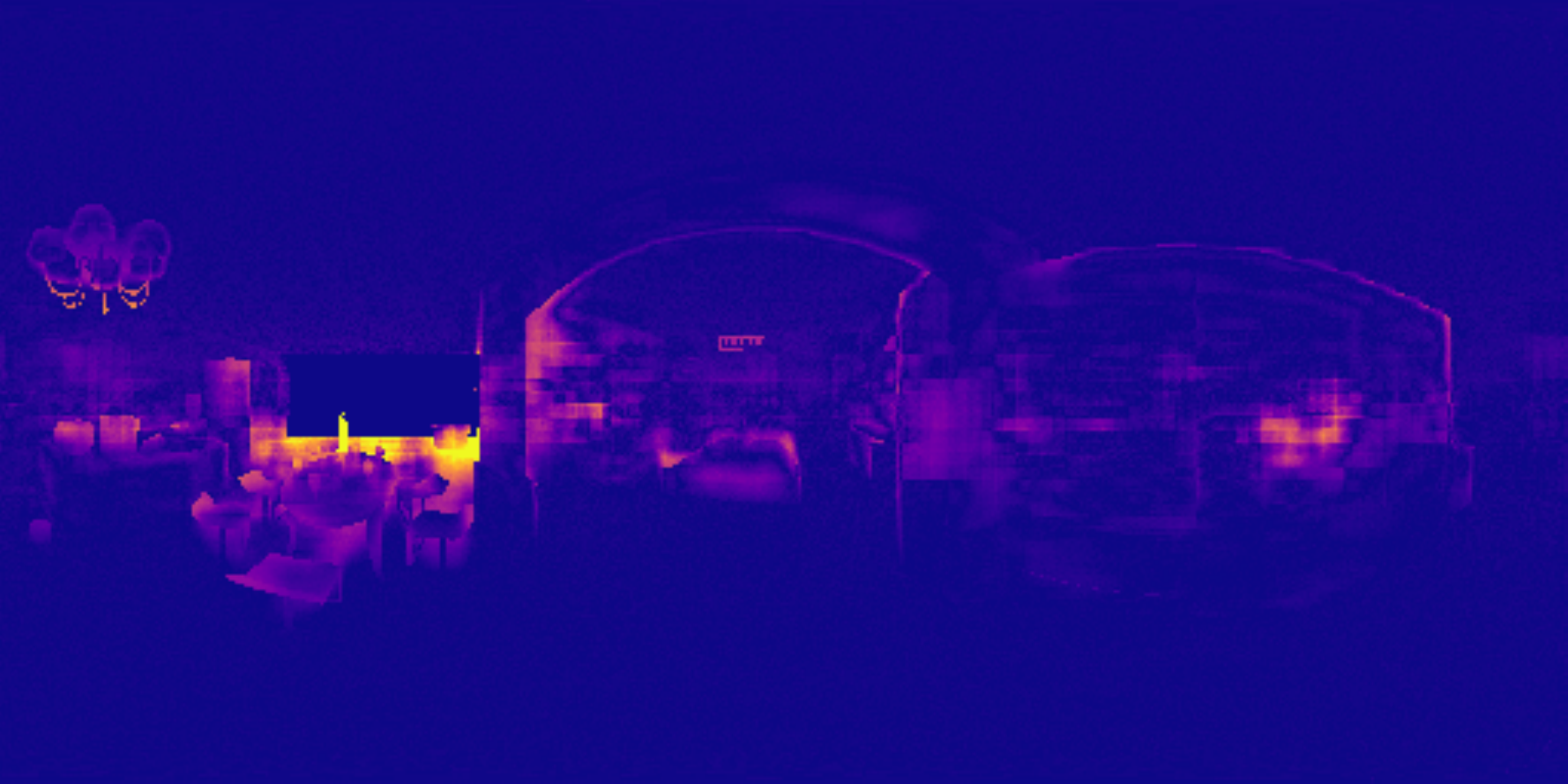}\put(2,40){\textcolor{white}{\textsf{Inv. Gnomonic \cite{coors2018spherenet,tateno2018distortion}}}}\end{overpic}
        \end{subfigure}
        ~
        \centering
        \begin{subfigure}[b]{0.32\textwidth}
            \begin{overpic}[width=\textwidth]{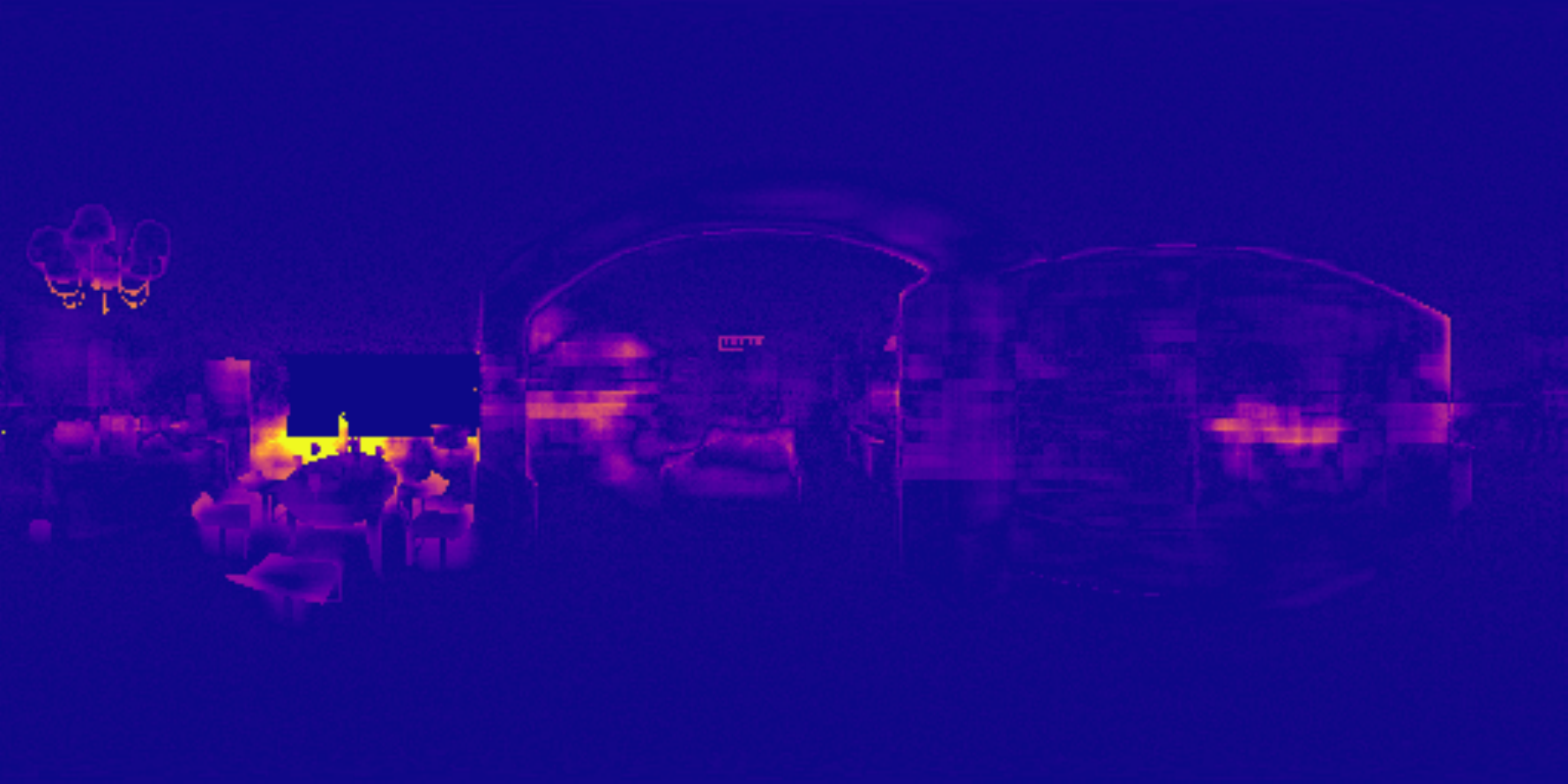}\put(2,40){\textcolor{white}{\textsf{Inv. Equirect. (ours)}}}\end{overpic}
        \end{subfigure}
        ~
        \centering
        \begin{subfigure}[b]{0.32\textwidth}
            \begin{overpic}[width=\textwidth]{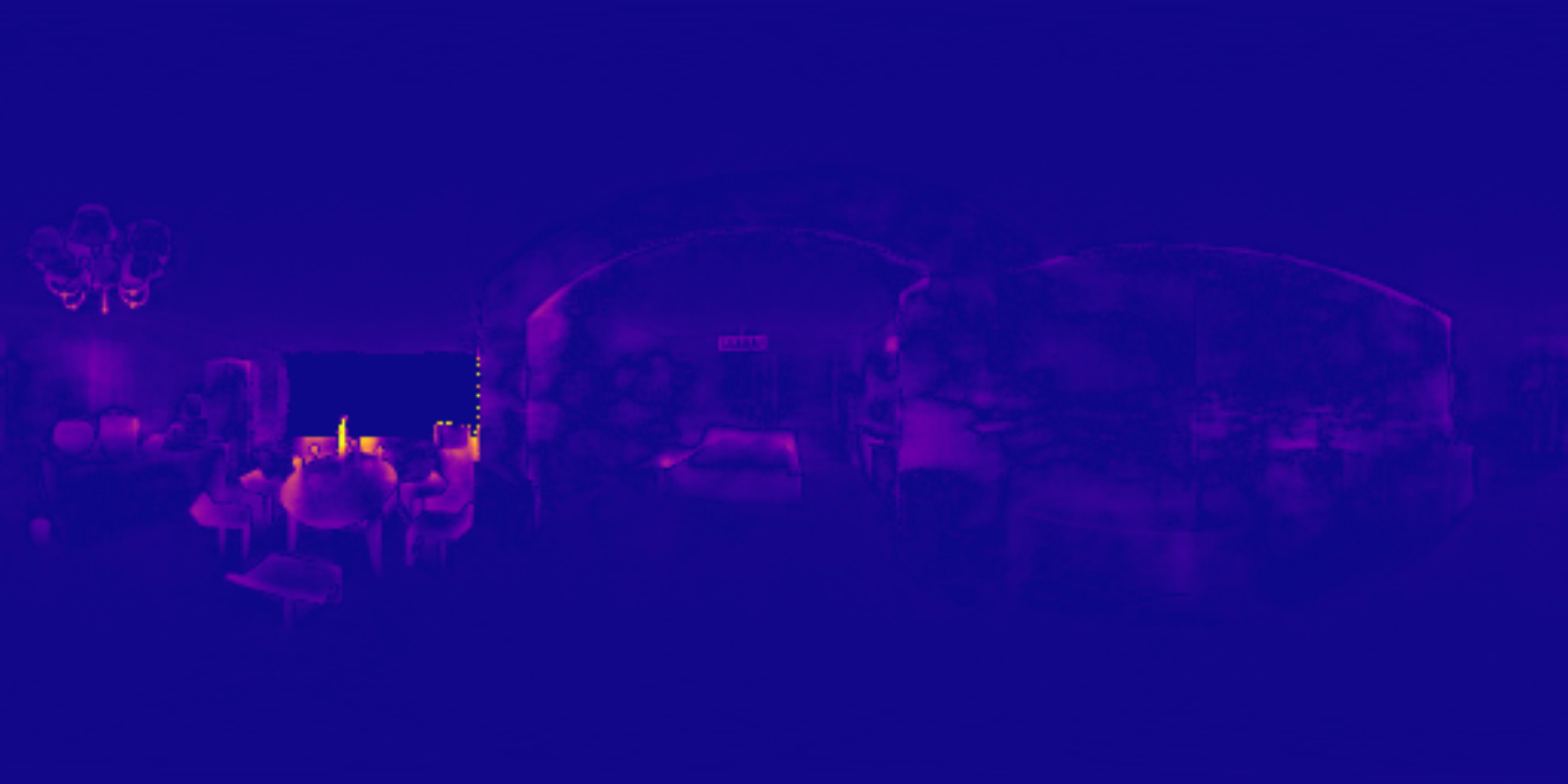}\put(2,40){\textcolor{white}{\textsf{ISEA (ours)}}}\end{overpic}
        \end{subfigure}
    \end{minipage}
    \hspace{-6mm}
    \begin{minipage}{.05\linewidth}
        \centering
        \begin{subfigure}[t]{\linewidth}
            \includegraphics[angle=90, origin=c, height=2.3cm, trim={1.5cm 2.5cm 1.5cm 0cm}]{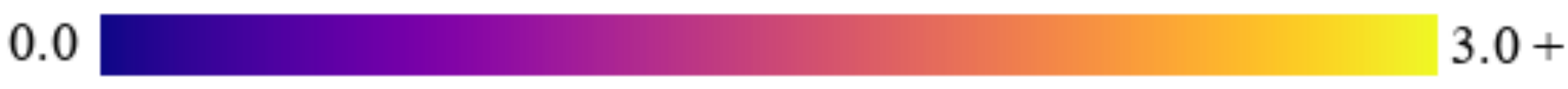}
        \end{subfigure}
    \end{minipage}\\
\vspace{-2mm}
\caption{Absolute error maps for depth predictions from an example equirectangular input image (top left). Our ISEA mapping function, enabled by our mapped convolution, results in visibly less error near the middle of the image. This is because the competing methods oversample the image poles, biasing the network against information near the equator.}
\label{fig:examples}
\vspace{-2mm}
\end{figure*}

\subsection{Cube maps}\label{sec:cubemaps}
We next analyze the cube map representation of spherical images. Cube maps are commonly used in graphics for texturing meshes as they allow for a low-poly representation of far away scenes. More importantly for spherical image inference, they are also a common format for large-scale $360^\circ$ image datasets: SUMO \cite{sumo} and Matterport3D \cite{Matterport3D} provide $360^\circ$ images in the cube map format.  Given their prevalence, we examine this representation as an input domain for dense inference.

Our results in the previous section suggest that we ought to be able to adapt to the distortion of any type of image projection by simply applying the right mapping function. For cube map projections of the sphere, the inverse function, from cube face coordinates $(u,v)$ to latitude and longitude $(\theta, \lambda)$, is given face-wise by Equation (\ref{eq:cubemapping}):
\begin{align}\label{eq:cubemapping}
\begin{array}{l|l|l}
Face & Latitude (\theta) & Longitude (\lambda)\\
\hline
\textbf{+X} & \tan^{-1}\left(\frac{v}{\sqrt{1+u^2}}\right) & \tan^{-1}(\frac{1}{-u})\\
\textbf{-X} & \tan^{-1}\left(\frac{v}{\sqrt{1+u^2}}\right) & \tan^{-1}(\frac{-1}{u})\\
\textbf{+Y} & \tan^{-1}\left(\frac{-1}{\sqrt{u^2+v^2}}\right) & \tan^{-1}(\frac{u}{v})\\
\textbf{-Y} & \tan^{-1}\left(\frac{1}{\sqrt{u^2+v^2}}\right) & \tan^{-1}(\frac{u}{v})\\
\textbf{+Z} & \tan^{-1}\left(\frac{v}{\sqrt{1+u^2}}\right) & \tan^{-1}(u)\\
\textbf{-Z} & \tan^{-1}\left(\frac{v}{\sqrt{1+u^2}}\right) & \tan^{-1}(u)\\
\end{array}
\end{align}
Sampling according to this function, we train our network on SUMO cube maps. For a fair comparison, we resize the cube maps to $128 \times 768$ to have an equivalent angular resolution to the equirectangular images we trained on above ($256 \times 512$). To see if the reduced pixel resolution affects learning, we also train on $256 \times 1536$ cube maps. 

\begin{figure}[ht]
\begin{center}
\includegraphics[width=1\linewidth]{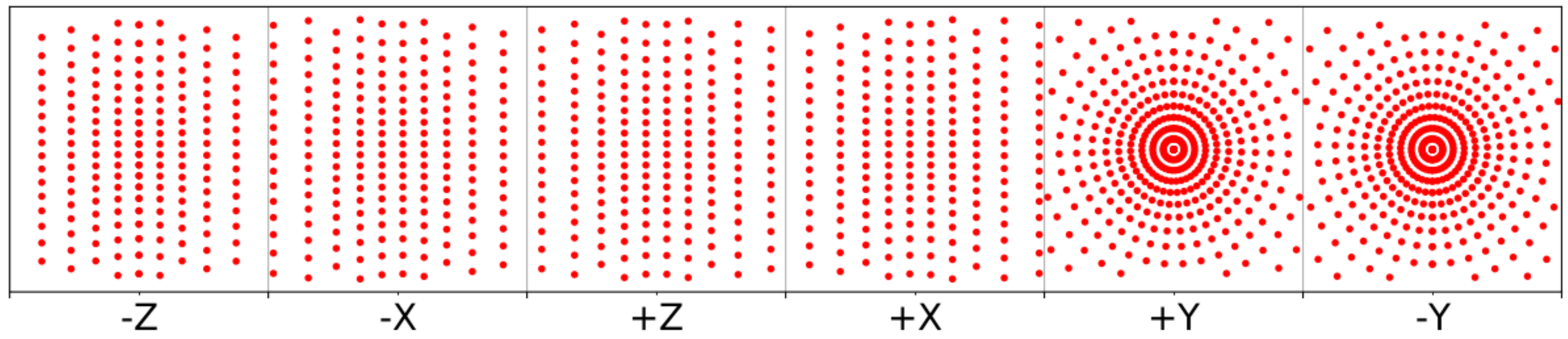}
\end{center}
\vspace{-4mm}
\caption{Sampling the cube map according to latitude and longitude. Notice the radial effect on the $\pm Y$ faces due to the poles of the sphere.}
\vspace{-4mm}
\label{fig:cubesampling}
\end{figure}

There is a simple mapping function between the two formats, so the choice between them is typically determined by the desired application. Despite this equivalence, we find that cube maps pose a challenge for CNNs, even when applying the inverse projection function. The results, shown in Experiment 2 in Table \ref{tab:mapprojresults}, are noticeably worse than our experiments on equirectangular images in the previous section. This suggests that choosing a mapping function according to the data domain is not enough. 

We posit that this worse result is due to the format's discretization. When we convolve on equirectangular images, we use a kernel whose grid is spaced according to a consistent angular resolution, the spacing between pixels. In general, the convolution operation expects that the filter and the input are discretized in the same way. Yet, cube maps are not uniformly indexed by latitude and longitude; they are indexed by a regular grid on the faces of the inscribing cube. Sampling according to spherical coordinates results in extreme radial transformations of the kernel on the $\pm Y$ faces due to the poles, as shown in Figure \ref{fig:cubesampling}, even though no such distortion is present there. The worse performance on the higher pixel resolution image is due to the increased disparity between the angular resolution defining the kernel spacing and the pixel resolution discretizing the image representation. We would expect this to have a larger effect as pixel resolution increases, which is borne out by our experiments. We conclude that although cube maps represent spherical content, they must be sampled according to their discretization of the sphere, rather than a measure on the sphere itself. While it may seem obvious, this is an important consideration when processing non-Euclidean domains: the data discretization matters in addition to the content represented.

\begin{figure}[ht]
\begin{center}
\includegraphics[width=0.6\linewidth]{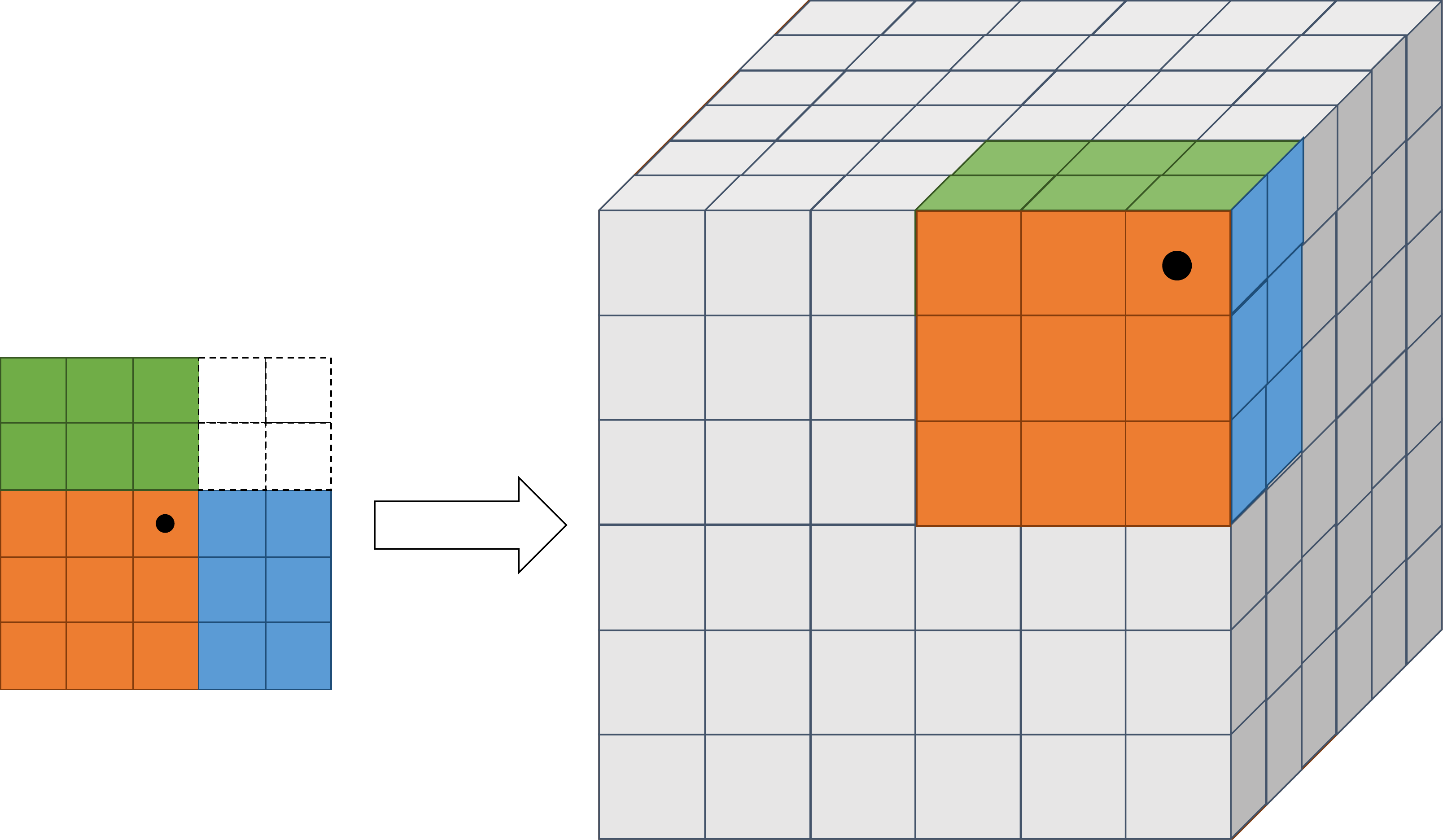}
\end{center}
\vspace{-2mm}
\caption{Ambiguity in applying a grid kernel (left) to the corner of a cube map (right). The upper-right $2 \times 2$ elements are lost in the fold. They would either doubly sample from the blue locations or the green locations.}
\label{fig:cubemapcorners}
\end{figure}

This concept therefore makes cube maps a difficult domain for CNNs. Should we use a mapping function that instead indexes the cube according to the pixel grid, we run into singularities on the corners, illustrated in Figure \ref{fig:cubemapcorners}. Furthermore, should we use a function that correctly maps the rectilinear distortion of each face, we lose the orientation of the sphere, which leads to ambiguous filter rotations. 
\subsection{Geodesic grids}\label{sec:icosphere}
Lastly, we demonstrate the application of mapped convolutions to 3D meshes in suggesting a new approach for spherical image inference based on a mapping to 3D icosahedral mesh. Our method provides an improved method for handling spherical distortion and resolves the previously unaddressed problem of information imbalance in planar spherical projections. In an equirectangular image, a pixel near the equator carries substantially more information than a pixel near a pole. In fact, pixels in the top and bottom rows are nearly redundant to their horizontal neighbors. This implies that points in different rows have different entropy when sampled by the convolutional kernel. Previous work has addressed spherical distortion patterns, but the existing literature has not yet confronted this concern. 

The icosahedral Snyder equal-area (ISEA) map projection \cite{snyder1992equal} handles this variation by mapping the sphere to a high-resolution geodesic polyhedron, specifically an icosahedral approximation to the sphere. This representation is nearly equal area and is largely isotropic. These two properties mean that information is close to uniformly distributed on the surface with little distortion. While there is no perfect projection of the sphere onto a planar surface, the ISEA projection is one of the least distorted \cite{kimerling1999comparing}.

\begin{figure}[ht]
\begin{center}
\includegraphics[width=\linewidth]{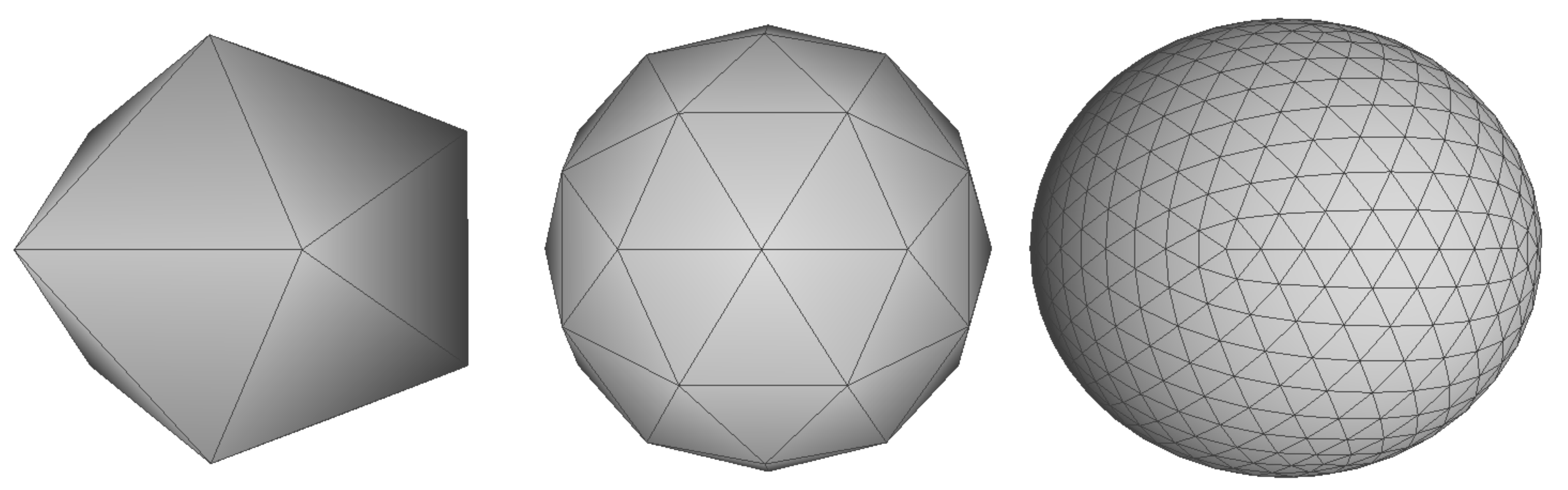}
\end{center}
\vspace{-2mm}
\caption{Subdividing an icosahedron to an icosphere.}
\vspace{-2mm}
\label{fig:icosphere}
\end{figure}

To use the ISEA projection, we create a $7^{th}$ order ``icosphere,'' a regular icosahedron subdivided $7$ times using Loop subdivision \cite{loop1987smooth} with the vertices subsequently normalized to the unit sphere. A visualization of this process is shown in Figure \ref{fig:icosphere}. We choose a $7^{th}$ order icosphere because the number of vertices  ($163,842$) is most similar to the number of pixels ($131,072$) in the $256 \times 512$ equirectangular images we use for comparison. 

Loop subdivision provides the useful property that the dimensions of the 3D mesh (the number of faces and vertices) roughly scale by 4 at each successive order. This is useful for maintaining the existing CNN paradigm, as typical stride-2 convolutions downsample the image by a factor of 4. For faces, the scale relation is exactly analogous, while for vertices, it is very close. For $|F|$ faces and $|V|$ vertices at subdivision $k$:
\begin{align*}
&|F| = 20(4^k) \\
&|V| = 
    \begin{cases} 
        12 & k = 0 \\
        12(4^k) - \sum_{i=0}^{k-1}6(4^i) & k > 0
    \end{cases}
\end{align*}

To use the ISEA projection, we map the spherical image to vertices of the icosphere. Our mapping function is: 
\vspace{-1mm}
\begin{equation}
M: (\phi, \lambda) \rightarrow F
\end{equation}
where $(\phi, \lambda)$ are the latitude and longitude of each pixel in an equirectangular image and $F$ is a face on the icosphere. We use barycentric interpolation on each face to assign pixel intensities to the vertices. We then define the mapped convolution operation according to the principles we derived in the previous two sections. We convolve on the icosphere using the inverse gnomonic projection given in Equation (\ref{eq:invgnomonic}). This choice is appropriate in this case, as we are mapping from a near-spherical mesh onto a tangent plane. Additionally, as the data is discretized by mesh vertices we apply the filter at each vertex with the filter resolution defined by the distance between adjacent vertices. During training, we compute the loss over each vertex rather than each pixel.

For an equivalent comparison with the prior art, we map our icosphere depth predictions back to equirectangular images. These results are listed under Experiment 3 in Table \ref{tab:mapprojresults}. Our ISEA mapping dramatically improves the results. The absolute error we report represents a 14\% improvement over our inverse equirectangular mapping proposed in Section \ref{sec:mapproj}, and a nearly 17\% improvement over the existing state-of-the-art method, inverse gnomonic sampling, from \cite{coors2018spherenet, tateno2018distortion}. Absolute error heat maps for each approach are shown in Figure \ref{fig:examples}. This visualization clearly shows the benefits of our proposed ISEA method. The use of the ISEA projection is made possible by our mapped convolution operation, which allows us to convolve on the surface of the icosphere. Without mapped convolution, we would be limited to more distorted map projection options and unable to resolve the information imbalance in this way.

\subsection{Benchmarking}
It is important to note that modifying the sampling function has a non-negligible effect on performance. Standard convolutions are efficient in part due to data locality: grid sampling benefits significantly from cache coherence. Mapped convolutions often break this locality and therefore see efficiency degrade as the input size increases. Additionally, our mapped \textit{col2im} function requires an atomic operation to handle real-valued sampling locations.
\begin{figure}[ht]
\begin{center}
\includegraphics[width=1.0\linewidth, trim={0.5cm 0cm 0.5cm 1.5cm}, clip]{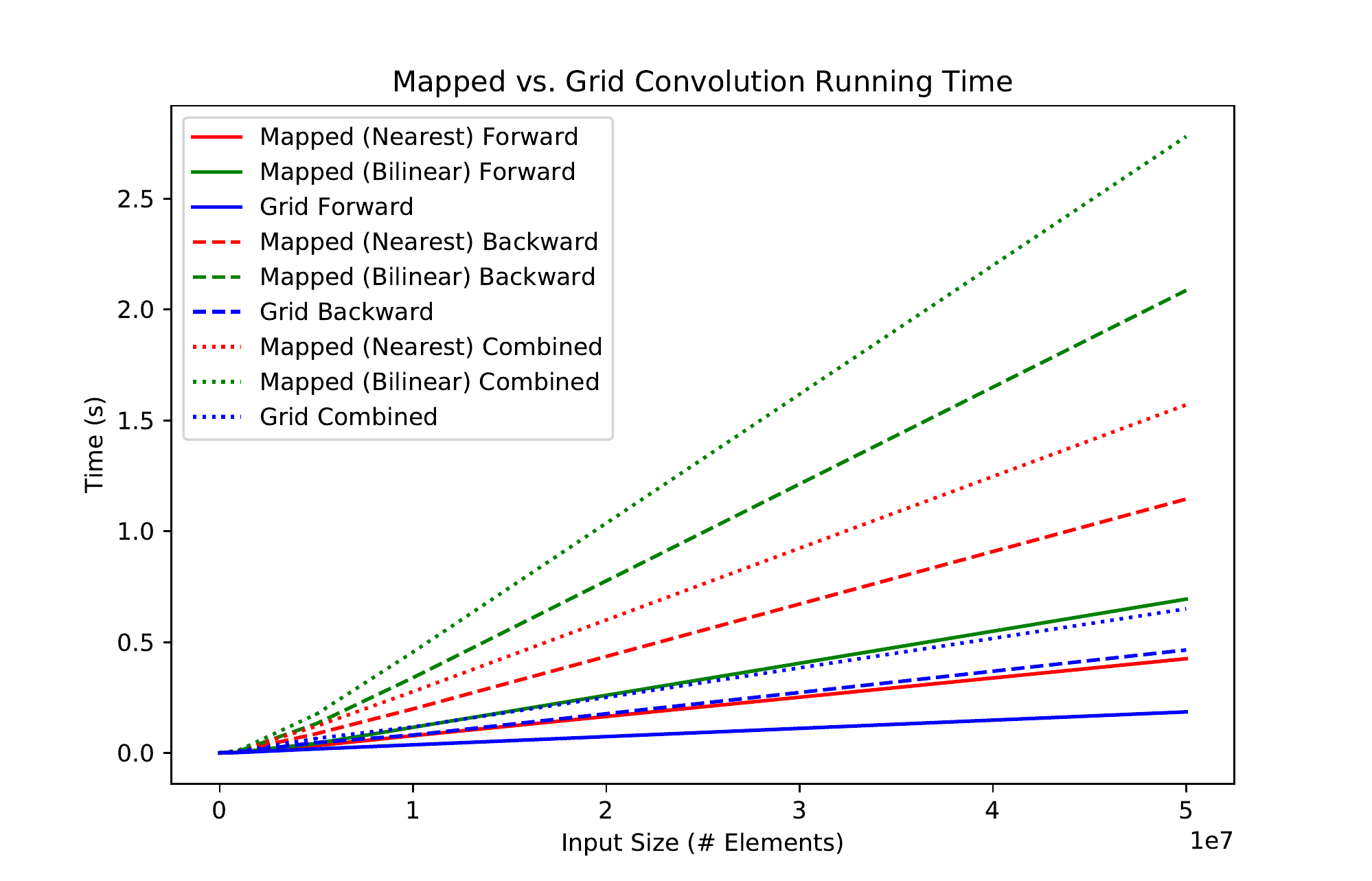}
\end{center}
\vspace{-5mm}
\caption{(Best viewed in color) Performance comparison between the mapped convolution and grid convolution operations on a single NVIDIA GeForce GTX 1080Ti GPU. We show results for a forward pass, a backward pass, and the combined forward and backward pass.}
\label{fig:performancegraph}
\vspace{-5mm}
\end{figure}
Hence, running time is now also tied to the choice of mapping function; for example, mapping functions where a single input location maps to a significant number of output location can inhibit training speeds by slowing down back-propagation. Even so, we find that mapped convolutions measure up to the performance needs of our applications. 

We profile our mapped convolution against the standard grid convolution implemented using the \textit{im2col} algorithm with double-precision floating point inputs. We compare to our own grid convolution implementation as popular frameworks like PyTorch and Tensorflow use additional optimizations to improve performance that are unrelated to the algorithm itself. The results are shown in Figure \ref{fig:performancegraph}. We run 100 trials at each input size and report the average result for each. For the mapping function, we shuffle pixels with uniform randomness. The breakdown of data locality clearly affects the speed of a forward pass as the images get larger, and the effect of the atomic adds in the backward pass is noticeable as well, but the slowdown is a constant factor. In this benchmark, the slowdown in the forward pass is $2.3\times$ for nearest-neighbor interpolation and $3.75\times$ for bilinear interpolation. For backward passes, the slowdown is $2.5\times$ and $4.5\times$ for nearest-neighbor and bilinear interpolation, respectively. It is useful to note that the largest input size profiled is equivalent to a 10-channel, $2000 \times 2500$ image; significantly larger than typical CNN inputs. While mapped convolutions are computationally costlier than grid convolutions, the impact is insignificant enough to not handicap usage. We have released our code as a PyTorch extension\footnote{\url{https://github.com/meder411/MappedConvolutions}}.

\section{Conclusion}
We have presented mapped convolutions, a versatile generalization of the convolution operation that divorces the data sampling from the weighted summation. Through the lens of spherical image depth estimation, we analyzed important details of designing an appropriate mapping function for an input domain. Drawing conclusions from our examination of equirectangular images and cube maps, we presented a new mapping function that processes the spherical image on the surface of an icosphere and results in a 17\% improvement over the state-of-the-art for dense spherical depth estimation. Moving forward, we see an opportunity for mapped convolutions to facilitate the application of CNNs to a host of new domains.

{\small
\bibliographystyle{ieee}

}
\end{document}